\documentclass[conference]{IEEEtran}
\IEEEoverridecommandlockouts

\usepackage{booktabs}
\usepackage{multirow}
\usepackage{makecell}

\usepackage{amsfonts}
\usepackage{amsmath}
\usepackage{subfigure}
\usepackage{diagbox}
\usepackage{tikz}
\usepackage[ruled,vlined,linesnumbered]{algorithm2e}

\usepackage{cite}
\usepackage{amsmath,amssymb,amsfonts}
\usepackage{algorithmic}
\usepackage{graphicx}
\usepackage{textcomp}
\usepackage{xcolor}
\def\BibTeX{{\rm B\kern-.05em{\sc i\kern-.025em b}\kern-.08em
    T\kern-.1667em\lower.7ex\hbox{E}\kern-.125emX}}
\begin{document}

\title{Edge Computing based Human-Robot Cognitive Fusion: A Medical Case Study in the Autism Spectrum Disorder Therapy
}

\author{\IEEEauthorblockN{Qin Yang}
\IEEEauthorblockA{\textit{Intelligent Social Systems and Swarm Robotics Lab (IS$^3$R)} \\
\textit{Computer Science and Information Systems Department}\\
Bradley University, Peoria, USA \\
email: is3rlab@gmail.com}
}

\maketitle

\begin{abstract}

In recent years, edge computing has served as a paradigm that enables many future technologies like AI, Robotics, IoT, and high-speed wireless sensor networks (like 5G) by connecting cloud computing facilities and services to the end users. Especially in medical and healthcare applications, it provides remote patient monitoring and increases voluminous multimedia.
From the robotics angle, robot-assisted therapy (RAT) is an active-assistive robotic technology in rehabilitation robotics, attracting researchers to study and benefit people with disability like autism spectrum disorder (ASD) children.
However, the main challenge of RAT is that the model capable of detecting the affective states of ASD people exists and can recall individual preferences. Moreover, involving expert diagnosis and recommendations to guide robots in updating the therapy approach to adapt to different statuses and scenarios is a crucial part of the ASD therapy process. This paper proposes the architecture of edge cognitive computing by combining human experts and assisted robots collaborating in the same framework to achieve a seamless remote diagnosis, round-the-clock symptom monitoring, emergency warning, therapy alteration, and advanced assistance. 



\end{abstract}

\section{Introduction}

Edge Computing offers the full computation or part of the computation that can process the data at the edge network, which enables low latency, faster response, and more comprehensive data analysis \cite{khan2019edge}. The connected devices can provide services in AI, robotics, autonomous driving, smart cities, healthcare, medical diagnosis, smart grids, multimedia, and security through the edge network. As one of the significant components of the smart city, the field of smart healthcare emerges from the need to improve the management of the healthcare sector, better utilize its resources, and reduce its cost while maintaining or even enhancing its quality level \cite{oueida2018edge}. Traditional smart healthcare systems can be divided into three layers: the collection layer (gathering the sensing data from patients), the transmission layer (sending the data to the base station through the intelligent terminal), and the analysis layer (analyzing the data in the cloud data center), which lack of real-time monitor, emergency service, and comprehensive disease analysis, high communication latency, inflexible network resource deployment, etc \cite{chen2018edge}. Although the recent 5G network can support edge computing–based healthcare systems, several challenges still hinder its application and benefits to the entire human community, such as large-scale healthcare, big data management, and patient information privacy \cite{hartmann2022edge}.
\begin{figure}[t]
    \centering
    \includegraphics[width=\linewidth]{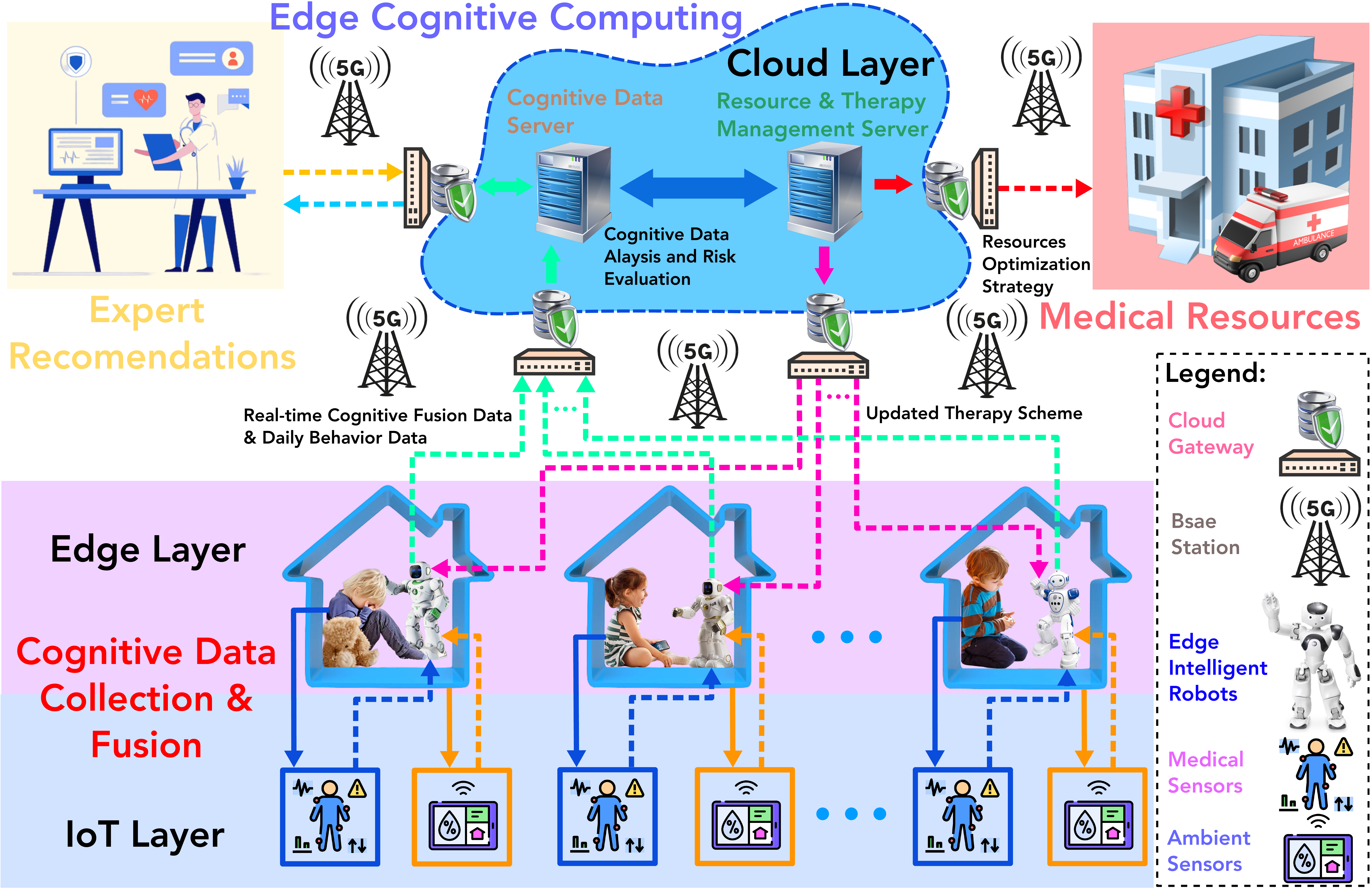}
    \caption{The illustration of the proposed architecture of cognitive computing based on Human-Robot cognitive fusion.}
    \label{fig:cognitive_computing_arch}
\vspace{-3mm}
\end{figure}
Especially when integrating AI and robotics technology, future medical and healthcare applications will involve vast amounts of real-time clinical data computation and analysis. Moreover, combining human experts' feedback based on real-time results from intelligent terminals is also critical for the future of medical systems and smart healthcare, which can provide professional recommendations, personalized services, and corresponding accurate medical treatment measures. Taking robot-assisted therapy (RAT) for autism spectrum disorder (ASD)\footnote{ASD is a developmental disability caused by differences in the brain, which affects the patients' normal interactions, such as learning, moving, communicating, and paying attention.} as an example, it is essential to enable RAT to adapt to individual unique and changing needs.
\cite{clabaugh2019long} formalized personalization as a hierarchical human-robot learning framework consisting of five controllers mediated by a meta-controller that utilized reinforcement learning to personalize instruction challenge levels and robot feedback based on each user’s unique learning patterns. To some extent, although it achieves long-term in-home deployments with children with ASD, it is hard to clarify the role of human experts in the therapy section and further improve their social skills in the real world. 
Therefore, the quality of the machine-to-machine communication and the human-robot interaction data fusion is the pre-condition for providing efficient and effective medical service, which can improve patient care experience and increase flexibility and adaptability from a business perspective \cite{wan2020cognitive}. 

From the cognitive computing perspective, \cite{chen2018edge} introduces the Edge-Cognitive-Computing-based (ECC-based) smart healthcare system, which can monitor and analyze users' physical health using cognitive computing and solve the problems of inflexible network resource deployment. \cite{yvanoff2020edge} develops a multi-language robot interface based on edge computing, helping evaluate seniors' mental health by interacting through questions. However, from the edge intelligent devices (like robots) angle \cite{groshev2023edge}, there are still several open questions in RAT for healthcare, such as what the best roles for robots are in therapy, how to develop a general approach to integrate robots into interventions adapting various patients' needs and recognize their status, and who among individuals with the specific symptoms are best suited for this approach, especially in RAT for ASD \cite{diehl2012clinical}. 

From the system design perspective, edging computing involves complex and heterogeneous architecture, which makes it hard to build a general framework for every edging computing application ultimately \cite{krishnasamy2020edge}. Particularly, smart healthcare systems are dynamic, flexible, and complex systems with unpredictable behaviors and need to organize resources and personalize diverse services efficiently \cite{oueida2018edge}.

In order to address those challenges, this paper proposes the architecture of the edge computing by combining human experts and assisted robots collaborating in the same framework to help ASD patients with long-term support. By integrating the real-time computing and analysis of a new cognitive robotic model for ASD therapy based on cognitive-behavior therapy (CBT) \cite{beck2020cognitive} with the proposed architecture, it can achieve a seamless remote diagnosis, round-the-clock symptom monitoring, emergency warning, therapy alteration, and advanced assistance. Fig. \ref{fig:cognitive_computing_arch} outlines the proposed architecture of edge cognitive computing and we list the main contributions of this research as follow:

\begin{itemize}
    \item We introduce the edge intelligent robot integrating cognitive data from patients and related ambient information by building the corresponding network architecture to achieve timely human expert recommendations and real-time medical resource optimization strategies;
    \item We design four stages of robot-assisted therapy (RAT) based on cognitive models of humor development to help ASD patients gradually master different levels of social and communication skills.
\end{itemize}

\section{Background and Preliminaries}

This section briefly reviews \textit{cognitive robotics} and provide a brief background to the \textit{cognitive-behavior therapy} (CBT). When describing a specific method, we use the notations and relative definitions from the corresponding papers.

\subsection{Cognitive Robotics}


Cognitive robotics studies the mechanisms, architectures, and constraints that allow lifelong and open-ended improvement of perceptual, reasoning, planning, social, knowledge acquisition, and decision-making skills in embodied machines \cite{merrick2017value}. Since cognitive modeling uses symbolic coding schemes depicting the world, perception, action, and symbolic representation become the core issues in cognitive robotics \cite{yang2019self,yang2020hierarchical,yang2021can}. Related disciplines are not just artificial intelligence and robotics but also neuroscience, cognitive science, developmental psychology, sociology, and so on \cite{asada2009cognitive}. Especially building a value system to mimic the ``brain'' of an AI agent mapping behavioral responses for sensed external phenomena is the core component of cognitive robotics, which is also an emerging and specialized sub-field in neurorobotics, robotics, and artificial cognitive systems research \cite{yang2023understanding}. 

Here, the value measures an agent's effort to expend to obtain a reward or avoid punishment \cite{yang2020needs}. It is not hard-wired for an AI agent, even a biological entity, and the specific value system achieved through experience reflects an agent's subjective evaluation of the sensory space \cite{10.1145/3555776.3577642}. And the value mechanisms usually have been defined as the {\it expected values}, particularly in uncertain environments. Moreover, the innate value reflects an agent's subjective evaluation of the sensory space, but the acquired value is shaped through experience during its development.

According to the development of cognitive robotics, the existing value systems can mainly be classified into three categories. {\it Neuroanatomical Systems} discuss the explainable biologically inspired value systems design from neuroanatomy and physiology perspectives \cite{sporns2000plasticity}; {\it Neural Networks Systems} build more abstract models through mathematical approaches to mimic the agent's value systems; {\it Motivational Systems} consider the model that agents interact with environments to satisfy their innate values, and the typical mechanism is reinforcement learning (RL) \cite{yang2024bayesian,yang2023hierarchical,yang2022self}.

\subsection{Cognitive-Behavior Therapy}

Cognitive behavioral therapy (CBT) is a type of psychotherapeutic treatment that helps people learn how to identify and change destructive or disturbing thought patterns that negatively influence their behavior and emotions \cite{hofmann2012efficacy}. 
It is based on the \textit{cognitive model} \cite{ellis1962reason}, which hypothesizes that people’s emotions, behaviors, and physiology are influenced by their perception of events.
In all forms of CBT that are derived from Beck’s model \cite{beck1964thinking}, treatment is based on a cognitive formulation, beliefs, and behavioral strategies that characterize a specific disorder \cite{alford1997integrative}. 


CBT involves a wide range of strategies to help people overcome negative patterns, such as identifying negative thoughts, practicing new social skills, SMART goal-setting, stress problem-solving, and self-monitoring \cite{beck2020cognitive}. It can be used as a short-term treatment to help individuals learn to focus on present thoughts and beliefs (symptoms like addiction, anger issues, anxiety, depression, etc.) and improve mental health conditions (like chronic pain, insomnia, stress management, etc.) \cite{hofmann2012efficacy}.

\section{Approach Overview}
This section discusses the details of the proposed architecture of cognitive computing and cognitive models of the edge intelligent robot based on cognitive-behavior therapy (CBT) for ASD.

\subsection{Cognitive Computing Architecture}

To organize the medical resources and manage a real-time healthcare system, it is crucial to define cognitive computing modules in the network edge clearly and implement corresponding cognitive analysis of the users' physical health data and ambient information. It can lower latency and guarantee the delivery of reliable and latest patient information and analysis results to doctors or experts.

Our proposed edge cognitive computing architecture (Fig. \ref{fig:cognitive_computing_arch}) leverages information collection and recognition, data fusion and analysis, resource strategy optimization, human expert recommendations, and real-time therapy updating, providing high energy efficiency, low cost, and high user Quality of Experience (QoE). 

\subsubsection{Cloud Layer}

In the cloud layer, we consider two kinds of servers -- \textit{Cognitive Data Server} and \textit{Resource and Therapy Management Server} -- to analyze cognitive data, evaluate patients' risks, optimize resource strategies, and update therapy schemes. We introduce them as follow:

\textit{a. Cognitive Data Server}

The server collects the cognitive fusion data from all edge intelligent robots, including users' physical signals and daily behavior data, and related internal network information such as the network type, service data flow, communication quality, and other dynamic environmental parameters. Moreover, according to users' priority level, the cognitive data server will distribute corresponding network and medical resources based on patients' risk evaluation. Furthermore, doctors or experts will receive the results of analyzed cognitive data and provide professional suggestions about the specific user case, such as updated prescriptions and medical instructions. Combining the surveillance cognitive information of users, the dynamic network resource information, and professional recommendations from experts provides the maximum edge computing resources to users based on their disease risk levels and significantly improves the QoE and healing probability.

\textit{b. Resource and Therapy Management Server}

Through receiving the analysis results from the cognitive data server, this server will learn resources about edge cloud computing, network communication, and medicine. Then, it optimizes scheduling strategies and distributes corresponding resources in real-time. Moreover, the server sends the integrated resource data back to the cognitive data server, updating its database. Specifically, the server can achieve resource optimization and energy saving by implementing the computing unload, handover strategy, caching and delivery, and the corresponding intelligent algorithms \cite{chen2018edge}, which can meet various heterogeneous application requirements.
\begin{figure*}[t]
\centering
\subfigure[Entry Level: an Example of the Assisted Robot's Behavior Trees of ``Aladdin and the Magic Lamp".]{
\label{fig:bt_el}
\includegraphics[width=0.65\columnwidth]{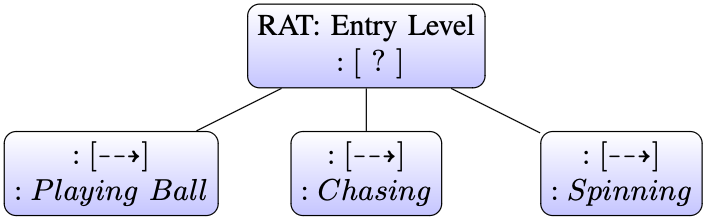}}
\subfigure[Basic Level: Assisted Robot's Behavior Trees]{
\label{fig:bt_bl}
\includegraphics[width=0.65\columnwidth]{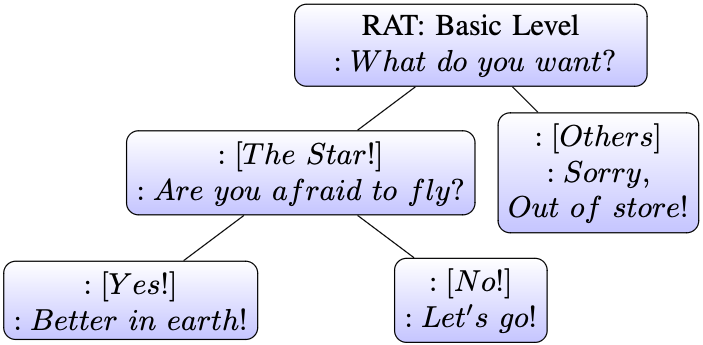}}
\subfigure[Middle Level: an Example of the Assisted Robot's Behavior Trees of the Knock knock joke.]{
\label{fig:bt_ml}
\includegraphics[width=0.65\columnwidth]{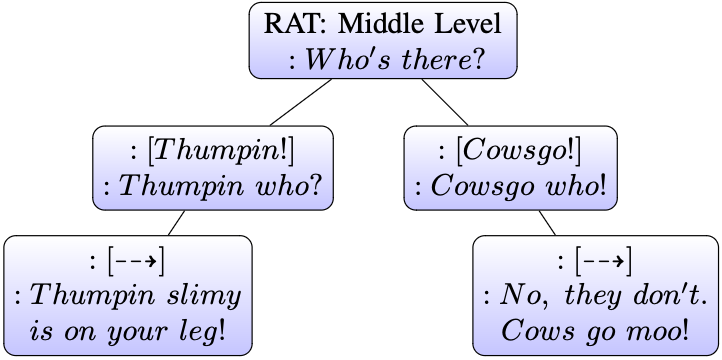}}
\vspace{-3mm}
\end{figure*}
\begin{figure}[t]
\centering
\subfigure[Advanced Level: Examples of Assisted Robot's Behavior Trees of the Sarcastic Jokes.]{
\label{fig:bt_al}
\includegraphics[width=0.6\columnwidth]{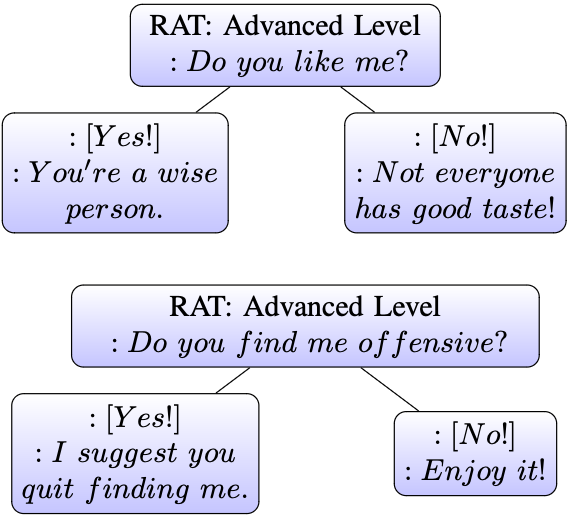}}
\vspace{-3mm}
\end{figure}
\subsubsection{Edge Layer}

Instead of constantly moving data to the cloud for computing operations, which accounts for the energy costs, in the edge layer, data can be mined and processed on edge devices and servers closer to the user \cite{bhargava2017fog}. Moreover, compared with traditional edge devices, we define a novel \textit{edge intelligent robot} for medical applications and smart health care, which not only can collect and process data from users and the environment but also serve as the medical equipment or tool to cure patients. 

Specifically, the edge intelligent robot needs to first integrate its perception data, such as voice, image, and video, with the data of medical sensors worn in patients and ambient sensors deployed in environments as a standard structure data flow uploading to the cloud layer through a high-speed sensor network like 5G. Furthermore, it executes the current or updated therapy scheme or instructions from the cloud layer to interact with patients, guiding them to fulfill the corresponding treatment. Moreover, it records all the interaction data and related medical parameters, sending them to the cloud layer for deeper analysis and human expert references.

More importantly, the edge intelligent robot can provide patients with 24-hour surveillance and care. If any emergency happens, it will inform the hospital or doctors through the network the first time and get the corresponding resources to tackle them, avoiding many human errors in traditional medical treatment.

\subsubsection{IoT Layer}

The IoT layer contains various devices and sensors, which can be classified into medical and ambient categories. The medical sensors were worn by patients, monitoring their status and recording related health parameters. They mainly collect the real-time physiological data of the user, which include electrocardiography (ECG), electromyography (EMG), respiration, heartbeat, body temperature, systolic pressure, and blood oxygen saturation (SpO2). The physiological data will be uploaded to the nearby edge computing node (edge intelligent robot) at the same time.

On the other hand, the ambient sensors are in charge of the surveillance of patient treatment environments and reserving various ambient information, such as ambient temperature, humidity, air quality, atmospheric pressure, etc., for further medical analysis and reference. Moreover, these devices transmit data to the edge intelligent robot and exchange information with the cloud through the high-speed communication network.

\subsection{Cognitive Models of the Edge Intelligent Robot}

In this section, we take the treatment of autism spectrum disorder (ASD) as an example and propose robot-assisted therapy (RAT) based on cognitive models of humor development to help ASD patients gradually master different levels of social and communication skills. 

Through above discussed edge cognitive computing network, the edge intelligent robot implements this therapy based on the results of cognitive data analysis and is guided by expert recommendations at each stage to cure the patients. We discuss more details as follow:

\subsubsection{Humor Styles for Cognitive Distortions}

The type of humor reflects the cognitive development of people and the level of their social skills \cite{bernet1993humor}. Humor styles are potential mediators of the association between cognitive and interpersonal vulnerability factors and psychological dysfunction, distress, or poor interpersonal functioning \cite{rnic2016cognitive}. To some extent, by improving their sense of humor, ASD patients can gradually enhance their communication skills and enter the mutual socialization process with others \cite{southam2005humor}.

Based on what McGhee \cite{mcghee1979humor} described, this research considers four stages of humor development in the whole CBT process. Using the assisted robot to design different funny scenarios and jokes treats ASD patients, especially the children, in their corresponding periods. Tab. \ref{table: humor_rat} illustrates the four stages of human development through RAT in our research.

\begin{table}[tbp]
\begin{center}
\caption{Four Stages of Humor Development through RAT}
\begin{tabular}{cccccc} 
\hline 
\makecell{\textbf{Therapy} \\ \textbf{Stage}} & \makecell{\textbf{Cognitive} \\ \textbf{Stage}} & \makecell{\textbf{Humor} \\ \textbf{Style}} & \textbf{RAT} & \makecell{\textbf{Data} \\ \textbf{Type}}\\
\hline
\makecell{Entry \\ Level}&\makecell{Sensorimotor \\ Stage}&\makecell{Incongruous \\ Actions}&\makecell{Funny \\ Behaviors}& images\\
\hline
\makecell{Basic \\ Level}&\makecell{Sensorimotor \\ Stage}&\makecell{Incongruous \\ Events}&\makecell{Interesting \\ Expression}&\makecell{Image, \\ Voice}\\
\hline
\makecell{Middle \\ Level}&\makecell{Preoperational \\ Stage}&\makecell{Conceptual \\ Incongruity}&\makecell{Knock- \\Knock Jokes}&\makecell{Image, \\ Voice}\\
\hline
\makecell{Advanced \\ Level}&\makecell{Concrete \\ Operations}&\makecell{Multiple \\ Meanings}&\makecell{Sarcastic \\ Jokes}&\makecell{Image, \\ Voice}\\
\hline
\end{tabular}
\label{table: humor_rat}
\end{center}
\vspace{-2mm}
\end{table}
\subsubsection{Behavior Tree based Humor Styles Representation}

According to the above discussion, we design four specific application cases representing them as behavior trees (BT) \cite{colledanchise2018behavior} to improve patients' experience and enhance the effect of the RAT in different stages of ASD.

\textit{a. Entry Level: Funny Behaviors}

Patients with less social skills enjoy interactions with familiar objects or games by creating something new or different out of them. Therefore, at the entry level, we design three scenarios (playing ball, chasing, and spinning) to let the assisted robot interact with patients, helping them develop interests and willingness to communicate and make new friends. Fig. \ref{fig:bt_el} shows the three scenarios as the BT.

\textit{b. Basic Level}

As patients develop interests and are not resistant to communicating with new partners, they become willing to use language and gestures for fun and to engage others. There may be some overlap between \textit{Entry Level} and \textit{Basic Level} in humor development, but the distinguishing feature of the \textit{Basic Level} is that a "verbal statement alone creates the incongruity and leads to laughter" \cite{mcghee1979humor}.

At this point, humor becomes a significant part of social skill development. In order to help patients gain positive emotional reactions from their partners, we create a simple pretend game\footnote{Pretend game is a loosely structured form of play that generally includes role-play, object substitution, and nonliteral behavior \cite{fein1981pretend}.} -- Aladdin and the Magic Lamp -- to let them play with the robot, adding social reinforcement to the fun. Fig. \ref{fig:bt_bl} illustrates an example of the Assisted Robot's Behavior Trees of ``Aladdin and the Magic Lamp". Here, the assisted robot plays the Magic Lamp and the patient plays the Aladdin.

\textit{c. Middle Level}

When patients master the above two levels of humor, their cognition will develop to Piaget’s Preoperational Stage \cite{papalia2007human}. Humor can be regarded as the intellectual play expressed through language \cite{freud1960jokes}, and the production and appreciation of humor also change. 

In this intellectual linguistic play, the listener needs to understand the double meanings of words to find it humorous \cite{southam2005humor}. At this stage, making others laugh becomes the reward of social approval, which can drive patients to develop this social language skill. Knock-knock jokes and other ready-made jokes are popular at this level. Considering knock-knock jokes and other ready-made jokes are popular, we design the scenarios of knock-knock jokes to support patients interacting with the robot at this level (Fig. \ref{fig:bt_ml}).

\textit{d. Advanced Level}

In the cognitive stage of concrete operations \cite{papalia2007human}, patients will improve mental operations to analyze multiple aspects of a situation and perform tasks at a higher level than they could in the preoperational stage. They need to understand the meanings of puns and other forms of more abstract humor and can use inductive and deductive reasoning and reversibility in thinking about the beginning, middle, and end points of a funny story or joke \cite{southam2005humor}. Teasing and sarcastic jokes are commonly used at this level. Fig. \ref{fig:bt_al} illustrates two examples of assisted robot's BT of sarcastic jokes.


\section{Conclusion and Future Works}

This paper introduces a novel architecture of edge cognitive computing integrating human experts and edge intelligent robots collaborating in the same framework to form the next generation of medical and smart healthcare systems. It can achieve a seamless remote diagnosis, round-the-clock symptom monitoring, emergency warning, therapy alteration, and advanced assistance.


For the next step, we want to implement our methods in real robots, such as Unitree Go2, and test them in the AWS wavelength framework. Furthermore, we want to apply the architecture in a real hospital medical system to test its robustness and effectiveness.

\section{Acknowledgments}
This work is supported by the NSF Foundational Research in Robotics (FRR) Award 2348013.


\bibliography{references}
\bibliographystyle{IEEEtran}

\end{document}